\documentclass[10pt,twocolumn,letterpaper]{article}

\usepackage[pagenumbers]{cvpr} 

\usepackage{amssymb,xcolor} 
\usepackage{tabularx}
\usepackage{graphicx}    
\usepackage[table]{xcolor} 

\newcommand{\cmark}{\textcolor{green}{\checkmark}}
\newcommand{\xmark}{\textcolor{red}{$\times$}}
\usepackage{booktabs}       
\usepackage{amsfonts}       
\usepackage{nicefrac}       
\usepackage{microtype}      
\usepackage{mathtools}
\usepackage{graphicx}
\usepackage{bm}
\usepackage{amsmath}

\def\eqref#1{equation~\ref{#1}}

\def\1{\bm{1}}

\def\mC{{\bm{C}}}

\def\mF{{\bm{F}}}

\def\mM{{\bm{M}}}

\def\mP{{\bm{P}}}

\def\mX{{\bm{X}}}

\def\mZ{{\bm{Z}}}

\DeclareMathAlphabet{\mathsfit}{\encodingdefault}{\sfdefault}{m}{sl}
\SetMathAlphabet{\mathsfit}{bold}{\encodingdefault}{\sfdefault}{bx}{n}

\definecolor{sectiongray}{gray}{0.93}
\usepackage{multirow}
\usepackage{siunitx}

\definecolor{cvprblue}{rgb}{0.21,0.49,0.74}
\usepackage[pagebackref,breaklinks,colorlinks,allcolors=cvprblue]{hyperref}

\def\paperID{21595} 
\def\confName{CVPR}
\def\confYear{2026}

\title{PETAR: Localized Findings Generation with Mask-Aware Vision-Language Modeling for PET Automated Reporting}

\author{
Danyal Maqbool$^{1,2}$, Changhee Lee$^2$, Zachary Huemann$^2$, Samuel D. Church$^{1,2}$, Matthew E. Larson$^2$, \\ Scott B. Perlman$^2$, Tomas A. Romero$^2$, Joshua D. Warner$^2$, Meghan Lubner$^2$, Xin Tie$^2$, \\
Jameson Merkow$^3$, Junjie Hu$^1$, Steve Y. Cho$^2$, Tyler J. Bradshaw$^2$\\
$^1$University of Wisconsin–Madison Department of Computer Sciences, Madison, WI, USA\\
$^2$University of Wisconsin–Madison Department Radiology, Madison, WI, USA\\
$^3$Microsoft, Redmond, WA, USA
}

\begin{document}
\maketitle
\begin{abstract}
Generating automated reports for 3D positron emission tomography (PET) is an important and challenging task in medical imaging. PET plays a vital role in oncology, but automating report generation is difficult due to the complexity of whole-body 3D volumes, the wide range of potential clinical findings, and the limited availability of annotated datasets. To address these challenges, we introduce PETARSeg-11K, the first large-scale, publicly available dataset that provides lesion-level correspondence between 3D PET/CT volumes and free-text radiological findings. It comprises 11,356 lesion descriptions paired with 3D segmentations. Second, we propose PETAR-4B, a 3D vision-language model designed for mask-aware, spatially grounded PET/CT reporting. PETAR-4B jointly encodes PET, CT, and 3D lesion segmentation masks, using a 3D focal prompt to capture fine-grained details of lesions that normally comprise less than 0.1\% of the volume. Evaluations using automated metrics show PETAR-4B substantially outperforming all 2D and 3D baselines. A human study involving five physicians---the first of its kind for automated PET reporting---confirms the model's clinical utility and establishes correlations between automated metrics and expert judgment. This work provides a foundational dataset and a novel architecture, advancing 3D medical vision-language understanding in PET.
\end{abstract}    
\section{Introduction}
\label{sec:intro}

Vision-language models (VLMs) have demonstrated considerable potential in automating radiology report generation, with promising applications for accelerating clinical workflows and mitigating radiologist burnout \cite{radburnout}. However, current research has predominantly focused on 2D imaging modalities—such as chest X-rays or individual computed tomography (CT) slices, while 3D modalities, such as CT, magnetic resonance (MRI), and positron emission tomography (PET) present greater technical challenges \cite{3DVLFM}. Of these, PET is severely underrepresented, despite its critical and expanding role in oncology for diagnosis, staging, and treatment response assessment. PET scans provide unique functional insights into metabolic activity, enabling early disease detection and treatment monitoring.

\begin{figure}[t]
  \centering
   \includegraphics[width=\linewidth]{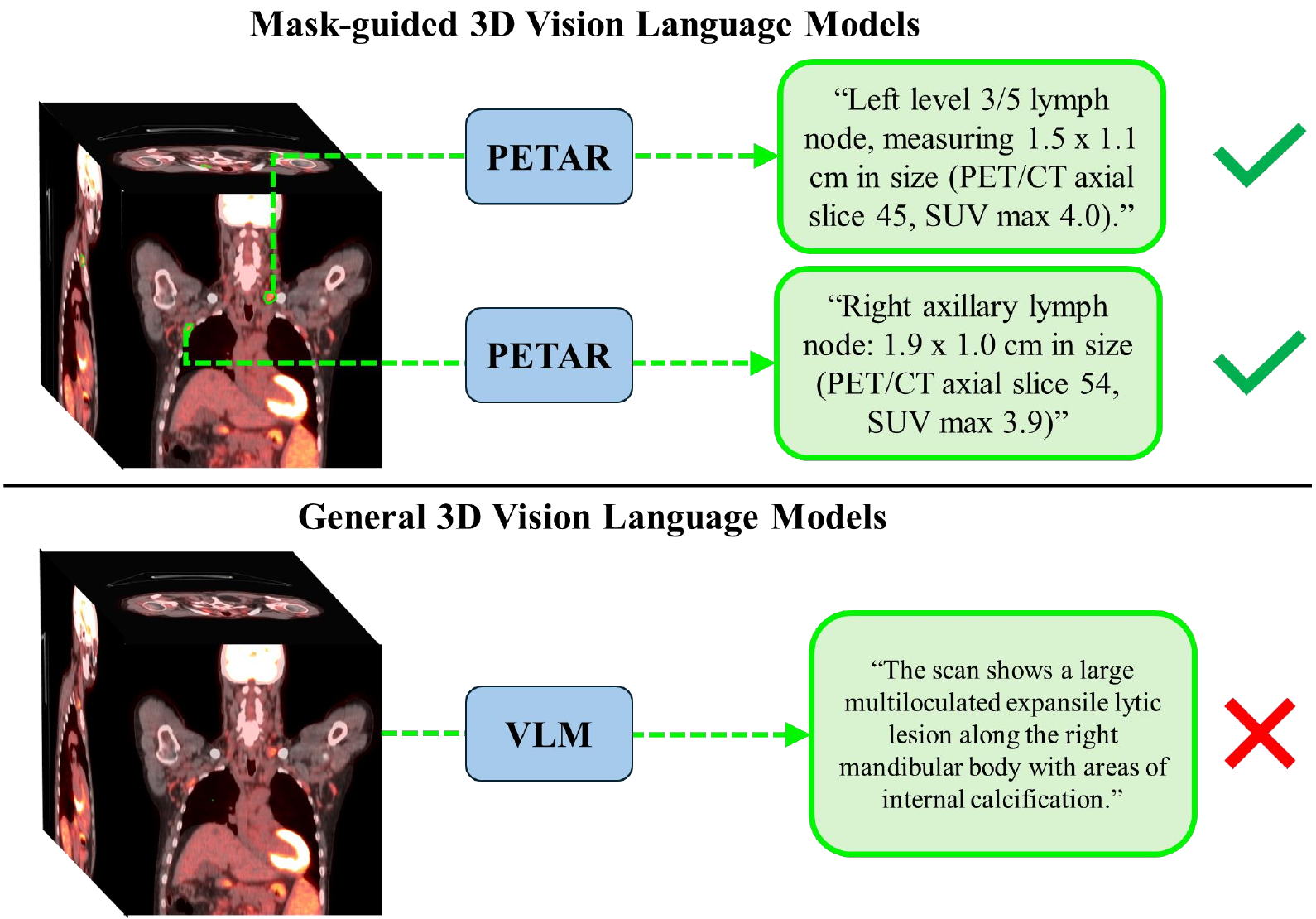}

   \caption{Overview of mask-guided PET/CT report generation. By incorporating lesion-level masks, PETAR produces anatomically fine-grained findings grounded in the 3D volume. In contrast, general 3D models perform global encoding without fine-grained anatomical correlation, hence they generate vague or clinically incomplete descriptions.}
   \label{fig:goal}
\end{figure}

PET presents several unique challenges that existing VLMs are poorly equipped to handle. First, clinical PET reports require fine-grained, lesion-level descriptions---not just global study summaries---which allows physicians to track individual findings over time. These descriptions are complex, as they often specify a lesion's anatomical site, sub-site, laterality, morphology, and metabolic activity, leading to a vast number of possible combinations (see \Cref{fig:goal}). This, together with the tendency to scan the entire body in a PET exam, makes PET reports arguably the longest in radiology, at times 3-fold longer than CT reports \cite{impressGen}. This combination of report length and descriptive granularity creates a significant challenge of fine-grained visual grounding. Second, these clinically relevant lesions can be numerous, very small (averaging less than 0.1\% of the total image volume), and spatially dispersed, making them challenging for standard vision encoders. Existing 3D VLMs, such as CT2Rep \cite{ct2rep}, M3D \cite{m3d}, and Merlin \cite{merlin} typically rely on global feature extraction and down-sampling, a process that can erase these fine-grained details \cite{reg2rg}. Furthermore, existing medical 3D VLMs have been predominantly trained on CT images (anatomical imaging) rather than PET images (molecular/metabolic imaging) and are not designed to jointly process dual-modality PET/CT volumes. Compounding this, a critical barrier is the lack of any large-scale, public dataset that provides the necessary 3D lesion-level segmentations aligned with free-text radiological findings. \Cref{fig:goal} illustrates our task and how conventional 3D medical VLMs fail to capture the specificity of lesion findings. Without accurate, lesion-specific findings, these models offer limited utility and reliability.

Our work addresses these issues by making two foundational contributions: a dataset and a model. Our dataset provides a direct connection between spatially localized lesion information and its corresponding natural language description, enabling precise alignment between imaging features and clinical language. Our model takes PET/CT scans and corresponding lesion segmentations as input to generate findings, leveraging the spatial cues in the masks to improve both specificity and interpretability. The segmentation masks can be obtained from advanced segmentation models \cite{autoPET} or FDA-cleared single-click PET lesion segmentation tools such as PetEdge \cite{PetEdge}. Our contributions are as follows:
\begin{itemize}
    \item We introduce \textbf{PETARSeg-11K}, a large-scale dataset of PET/CT studies with explicit lesion-level alignment between image space and report text. The dataset comprises \textbf{11,356 lesion descriptions} paired with corresponding 3D lesion segmentations, collected from over \textbf{5,000 whole-body PET/CT exams}. Each lesion is associated with a referring expression extracted from clinical radiology reports using a hybrid pipeline that couples anatomical cues with LLM-based parsing and normalization. To our knowledge, this is the first dataset to link localized PET/CT lesion masks with free-text descriptions.
    \item We propose \textbf{PETAR-4B}, a 3D mask-aware VLM designed for PET/CT findings generation. PETAR-4B jointly encodes PET, CT, and 3D lesion masks, enabling the model to condition language generation on both global disease context (e.g., disease distribution) and fine-grained lesion attributes (e.g., metabolic activity).
    \item We present a comprehensive evaluation of PETAR-4B and baseline 2D and 3D VLMs on lesion-level PET/CT findings generation. We employ automated evaluation metrics and human evaluations to quantify clinical usefulness. We benchmarked these metrics against expert physician preferences to understand which metrics are most reliable for PET findings generation.
\end{itemize}

\section{Related work}
\label{sec:related}

\subsection{Medical vision language models (VLMs)}
Medical VLMs encode multimodal inputs, such as medical images, into a shared embedding space, and leverage large language models to generate clinically meaningful outputs. MAIRA \cite{maira1, maira2} generates detailed radiology reports from chest X-rays through aligned vision–language encoders, while LLaVA-Med \cite{llavamed} adapts general-domain VLMs to biomedical applications via curriculum learning on figure–caption pairs. However, these methods are limited to the 2D plane, discarding critical spatial information inherent in 3D medical volumes.

To extend multimodal reasoning into volumetric contexts, M3D \cite{m3d} leverages a large-scale multimodal dataset for 3D segmentation and report generation. Merlin \cite{merlin} aligns 3D CT data with both structured and unstructured clinical information within a unified embedding space, and CT2Rep \cite{ct2rep} introduces a causal 3D feature extractor for automated chest CT report generation. RadFM \cite{radfm}, trained on both 2D and 3D data, seeks to build a generalist model capable of handling diverse radiologic tasks across modalities. A crucial gap within these works is that they are \textit{mask-agnostic}, limiting their ability to localize and describe lesions or regions with fine-grained precision. 

The datasets underpinning these models primarily rely on CT and X-ray modalities, with CT dominating recent 3D efforts. Several PET/CT datasets exist but each lacks key elements needed for lesion-level captioning. As summarized in \Cref{tab:dataset_comparison}, most public PET/CT datasets provide only images or masks, but lack the accompanying text findings needed for lesion-level captioning. ViMed-PET \cite{vietnam} and Pet2Rep \cite{pet2rep} include textual findings but lack grounded lesion masks. Together, these datasets capture only parts of the task, underscoring the need for a resource that combines findings and masks for training PET/CT VLMs.
\subsection{Fine-grained captioning}
Advances have been made toward fine-grained visual grounding in 2D images. Early methods, such as Kosmos \cite{kosmos}, Shikra \cite{shikra}, and GPT4ROI \cite{gpt4roi} incorporate explicit spatial cues—such as bounding box coordinates—into textual prompts to direct visual attention. More recent approaches have introduced flexible region-level interactions, allowing segmentation masks or point-based cues to guide the model’s focus and improve spatial grounding, such as AlphaClip, and ViP-LLaVA \cite{alphaclip, finecaption, vipllava, osprey}. To further improve fine-grained analysis, Describe Anything Model \cite{dam} introduces a zoomed in \textit{focal prompt} as input. This zoomed in input further improves fine-grained analysis capabilities by incorporating both global and local information.

In the medical domain, Reg2RG \cite{reg2rg} employs these ideas by introducing a mask input channel to generate more detailed, region-aware findings from CT scans. However, it is designed to use large, organ-level masks to generate region-level descriptions (e.g., entire lung). Its scope is also limited to single-modality chest CT and is not designed for the fine-grained abnormality-focused task of lesion-level captioning. Our work builds upon this direction by introducing a 3D, mask-aware architecture capable of jointly processing dual-modality PET and CT scans. Crucially, it uses mask-guided focal prompting to enable localized, clinically grounded captioning of small, discrete lesions across the body, a task that is anatomically and technically distinct.

\section{PETARSeg-11K dataset construction}
\label{sec:dataset}

\subsection{Preliminaries on PET/CT reports}
PET reports describe how radiotracers are distributed in the body in a PET scan. This is usually complemented by a CT or MRI scan for anatomical reference. Within each body region, reports list the lesions or regions with abnormal radiotracer uptake and include measurements such as standardized uptake value (SUV\textsubscript{max}), slice numbers, and qualitative descriptions. Because lesions are usually tracked across multiple scans, this information helps future readers identify the lesions of concern. 


\subsection{Dataset construction pipeline}
This research was conducted under a protocol approved by our Institutional Review Board (IRB), which granted a waiver of informed consent. All data was fully anonymized.

\begin{figure}[t]
  \centering
   \includegraphics[width=\linewidth]{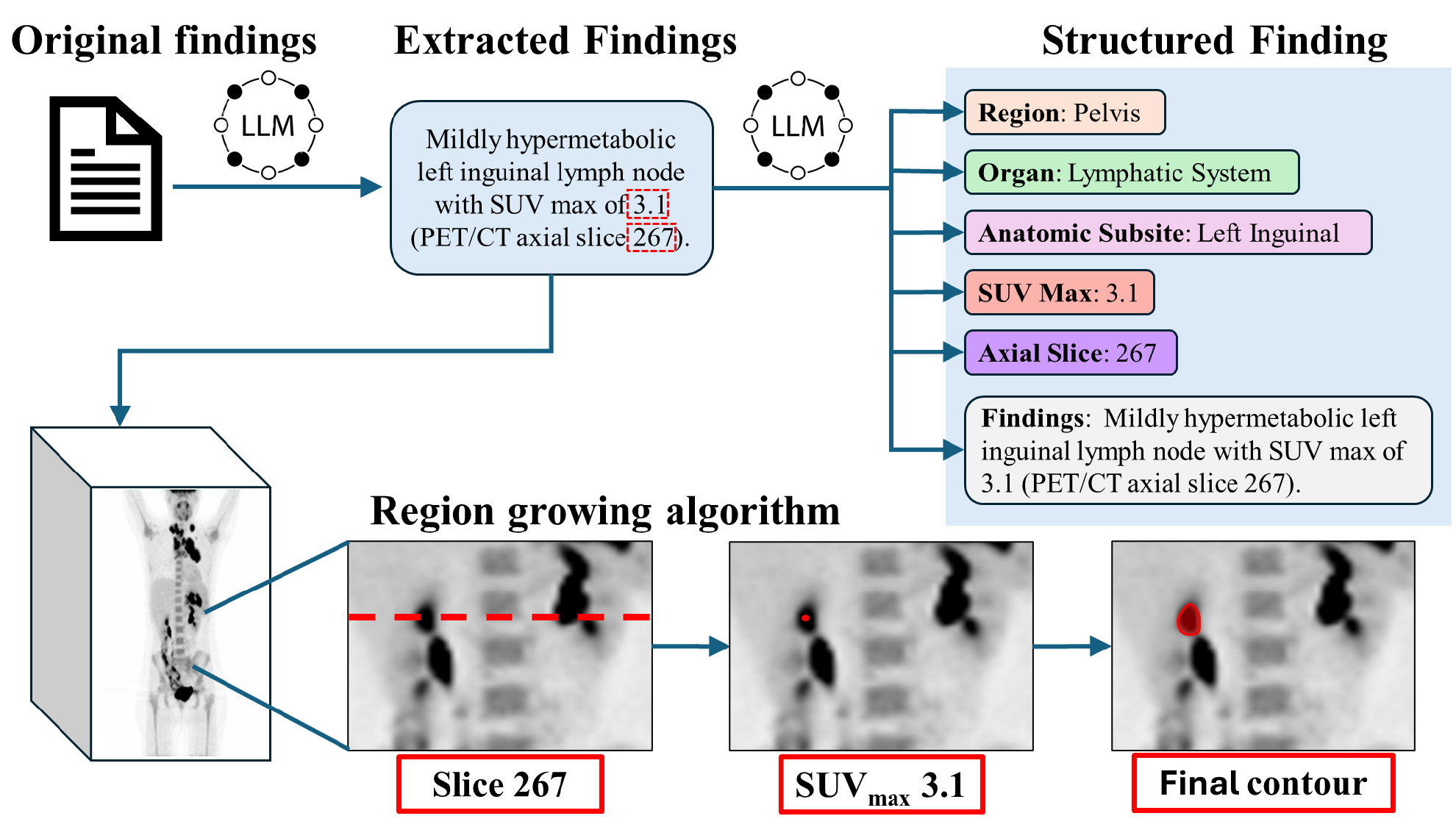}
   \caption{Overview of the PETARSeg-11K data pipeline. LLMs extract key lesion attributes (e.g., SUV\textsubscript{max} and slice number) from radiology reports, which guide a region-growing algorithm to localize and refine PET lesions. The final findings are structured into standardized fields linking text to 3D lesion masks.}
   \label{fig:data}
\end{figure}

\begin{table}[tbh]
    \centering
    \caption{Comparison of publicly available PET/CT datasets}
    \resizebox{\columnwidth}{!}{%
        \begin{tabular}{lcccccc}
            \toprule
            \textbf{Dataset} & \textbf{Total scans} & \textbf{Findings} & \textbf{Grounded Masks} & \textbf{Whole body} \\
            \midrule
            RIDER Lung PET-CT \cite{Muzi2015_RIDER_Lung_PET_CT} & 244 & \xmark & \xmark & \xmark \\
            Head-Neck PET-CT \cite{Vallieres2017_HeadNeckPETCT} & 298 & \xmark & \xmark & \xmark \\
            Lung-PET-CT-Dx \cite{Li2020_LungPETCTDx} & 251 & \xmark & \xmark & \xmark \\
            FDG-PET-CT-Lesions \cite{autoPET} & 1014 & \xmark & \cmark & \cmark \\
            SPADe \cite{spade} & 1614 & \xmark & \xmark & \cmark \\
            SegAnyPet \cite{zhang2025seganypet} & 5,731 & \xmark & \cmark & \cmark \\
            Pet2Rep \cite{pet2rep} & 565 & \cmark & \xmark & \cmark \\
            ViMed-PET \cite{vietnam} & 2757 & \cmark & \xmark & \cmark \\
            PETARSeg-11K (ours) & 5126 & \cmark & \cmark & \cmark \\
            \bottomrule
        \end{tabular}
    }
    \label{tab:dataset_comparison} 
\end{table}

We collected a set of 33,000 PET/CT scans from patient hospital records. Then, we leveraged the pipeline developed by Huemann \etal \cite{contextual} to create grounded segmentations for metabolic lesions. The method uses an ensemble of language models to filter out irrelevant sentences, disambiguate references to prior studies, and accurately extract the SUV\textsubscript{max} value and corresponding slice number. Specifically, we used an ensemble of Mistral-7B-Instruct and Mixtral-8×7B-Instruct. Lesion masks were generated using an iterative thresholding algorithm \cite{iterative} applied to the PET volume. The algorithm first applies a threshold based on the reported SUV\textsubscript{max}, producing an initial set of candidate regions. Connected components are identified, and the component whose SUV\textsubscript{max} matches the reported SUV\textsubscript{max} (within $\pm0.1$) and intersects the reported axial slice is selected. From the matching SUVmax and slice, we start at the max pixel and iteratively grow the contour until it stabilizes. The resulting dataset contains 11,356 lesion descriptions across 5,126 unique exams, covering multiple radiotracers including $^{18}$F-fluorodeoxyglucose (FDG), $^{68}$Ga-(DOTA-(Tyr$^3$)-octreotate) (DOTATATE), $^{18}$F-fluciclovine, and $^{18}$F-DCFPyL. The data was resampled to \SI{3}{mm} resolution with dimensions $192\times 192 \times 352$. Physician evaluations of a sample of this dataset recorded a 98\% accuracy in contour locations \cite{contextual}.

Using a strong LLM, Qwen3-30B-A3B \cite{qwen3}, each description is then formatted into a structured schema. An illustration of a complete sample is provided in \Cref{fig:data}. This representation explicitly grounds spatial and anatomical references.

To strengthen the model’s understanding of global anatomical structure, we curated a complementary pretraining dataset. Using TotalSegmentator \cite{TotalSegmentator}, we automatically segmented the CT volumes of our PET/CT dataset, generating roughly 100,000 labeled segmentations spanning its 117 predefined anatomical classes. This dataset provides comprehensive anatomical coverage, and encourages model alignment between metabolic activity in PET scans and the structural anatomies in the CT scan. 

Our dataset is the first publicly available multimodal PET/CT dataset with lesion-level granularity. As we highlight in \Cref{tab:dataset_comparison}, existing PET/CT datasets are typically limited in scope, focusing on specific organs, lacking text, lesion-level findings, or missing segmentation masks. In contrast, our dataset, which we term \textbf{PETARSeg-11K}, provides the first large-scale, whole-body PET/CT collection with aligned findings, 3D lesion masks, and textual descriptions. 

\begin{figure*}[tbh]
  \centering
  \includegraphics[width=\linewidth]{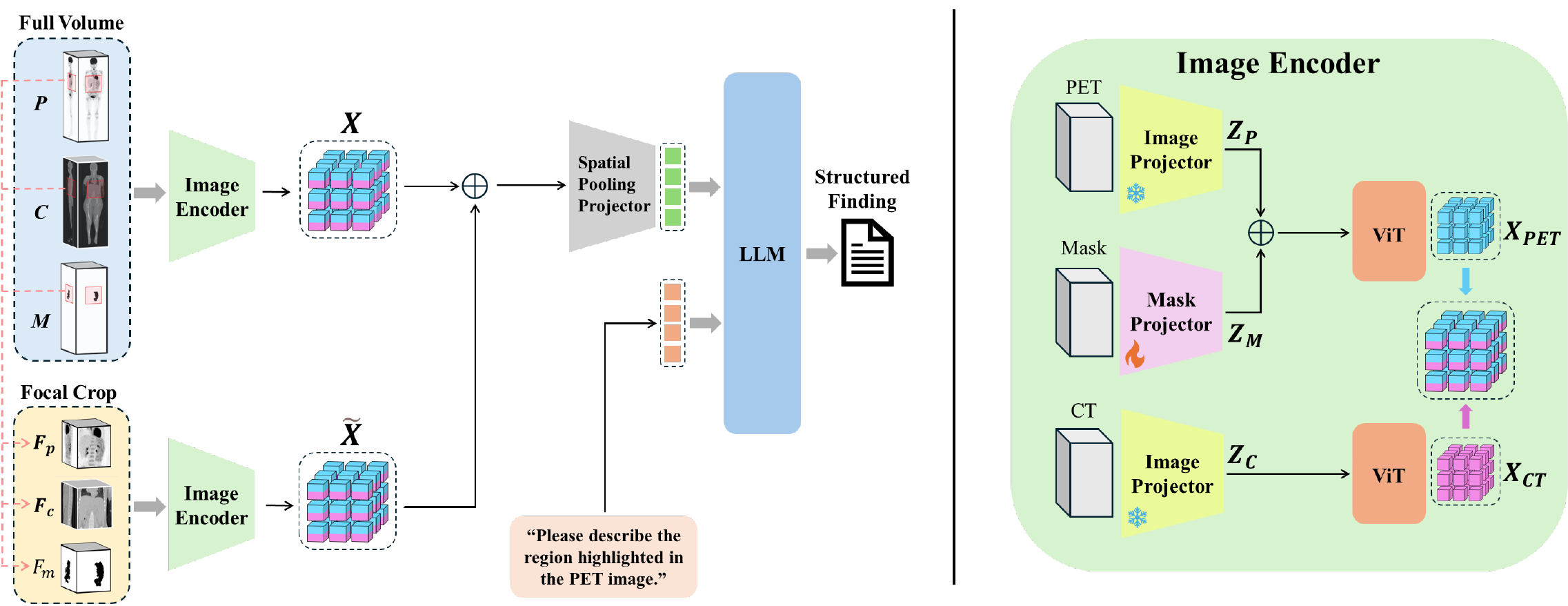}
  \caption{
  \textbf{Overview of the proposed framework.} 
  The \textbf{left panel} illustrates the overall architecture, which integrates PET, CT, and lesion mask inputs through 3D convolutional image projectors and an M3D-CLIP backbone. The resulting multi-modal visual tokens are fused and spatially pooled before being passed to a Phi3-4B language model that generates clinically grounded text descriptions conditioned on visual features and textual prompts. 
  The \textbf{right panel} details the \textit{Image Encoder} design, where modality-specific projectors (for PET, CT, and mask inputs) map each input into a shared latent space. These embeddings are subsequently processed by a ViT encoder to produce modality-aligned visual tokens for downstream fusion.
  }
  \label{fig:overall}
\end{figure*} 

\section{PETAR model design}
\label{sec:model}

\subsection{Mask aware inputs}
Given a 3D PET scan \( \mP \in \mathbb{R}^{D \times W \times H} \), a CT scan \( \mC \in \mathbb{R}^{D \times W \times H} \), and a binary mask \( \mM \in \{0,1\}^{D \times W \times H} \) indicating the region of interest, the objective is to generate detailed diagnostic findings $y$ focused on the masked region. Let \( f_{\theta} \) denote our model parameterized by \(\theta\). The generation is:
\begin{align}
    y = f_{\theta}(\mP, \mC, \mM).
\end{align}


\subsection{Focal prompt} \label{sec:focal_prompt}
A key bottleneck in lesion-level understanding is that lesions are often a very small part of a volume, and so standard approaches lead to a risk of information loss during common global processing steps, such as resizing, cropping, or downsampling. For instance, in our dataset, the average binary mask $\mM$,  occupies less than 0.1\% of the total scan volume, and the sizes range from 0.01\% to 2\%. To address this, we extended the ideas of the Describe Anything Model \cite{dam} to create a 3D focal prompt input to provide a localized, high-resolution view of the lesion.

We first center the crop on the mask $\mM$ and extract a cubic subvolume covering the region of interest. To enhance robustness and avoid overfitting to fixed spatial positions, we apply random spatial perturbations to both the cube center and its side length, limiting deviations to 20\% of the original mask center and cube size. The crop is adjusted such that the mask remains fully contained within the subcube, ensuring complete lesion visibility in the focal prompt.

Specifically, we denote the lesion mask center as \( c \in \mathbb{R}^3 \) and the desired crop size as \( r \).
We apply the small random perturbations to both the center and size as follows:
\begin{align}
    \tilde{c} = c + \triangle c, \quad \tilde{r} = r + \triangle r, \\
    \triangle c, \, \triangle r \stackrel{\text{i.i.d.}}{\sim} \mathcal{U}(-0.2r, 0.2r),
\end{align}
where \( \triangle c \) denotes a random translation offset, and \( \triangle r \) represents a random variation in crop size. 

The focal PET, CT, and mask crops are then extracted as:
\begin{align}
    \mF_P, \mF_C, \mF_M = \text{Crop}(\mP, \mC, \mM; \tilde{c}, \tilde{r}),
\end{align}
where \(\mF_P, \mF_C, \mF_M \in \mathbb{R}^{\tilde{r}\times\tilde{r}\times\tilde{r}}\) represent the PET, CT, and mask focal crops, and \(\text{Crop}(\cdot)\) denotes the operation that extracts the 3D subvolume. These focal crop inputs can be seen in \Cref{fig:overall}.

\subsection{Shared visual encoding} \label{sec:shared_visual_encoding}
For a more efficient architecture, we adopted a shared 3D vision transformer \cite{vit} to encode both PET and CT volumes. PET scans on their own contain limited structural information about the body, and are complemented by CT scans for anatomical context. Unlike prior work, we propose to jointly process them to capture both pieces of information. Specifically, to prepare the sequential inputs for the shared transformer, the PET and CT volumes are split into non-overlapping 3D patches of size $(s_D, s_W, s_H)$, which are then flattened into a sequence of $K$ patches and linearly projected into token embeddings of dimension $d$.
\begin{align} \label{eq:Z_P}
    \mZ_P = \text{PatchEmbed}(\mP) \in \mathbb{R}^{K\times d} \\
    \mZ_C = \text{PatchEmbed}(\mC) \in \mathbb{R}^{K\times d}
\end{align}
where $\mZ_C$ and $\mZ_P$ are the patch embeddings. The mask embeddings are produced using a different set of parameters.  
\begin{align} \label{eq:Z_M}
    \mZ_M = \text{PatchEmbed}(\mM) \in \mathbb{R}^{K\times d}
\end{align}

The 3D vision transformer, denoted by \(\mathcal{T}\), encodes the embeddings while incorporating the lesion mask into the PET scan via element-wise additive conditioning:
\begin{align} 
\mathbf{X}_{\text{PET}} = \mathcal{T}\big(\mathbf{Z}_P +\mathbf{Z}_M\big),~~~
\mathbf{X}_{\text{CT}} = \mathcal{T}\big(\mathbf{Z}_C\big).
\end{align}

We then fuse the CT and mask-aware PET token embeddings to obtain the global features $\mX$ by concatenating them along their embedding dimension:
\begin{align}\label{eq:X}
    \mathbf{X} = \mathrm{Concat}(\mathbf{X}_{\text{PET}}, \mathbf{X}_{\text{CT}}) \in \mathbb{R}^{K\times d_v}.
\end{align}

The same process from Eq.~(\ref{eq:Z_P})-(\ref{eq:X}) is applied to the focal crops \((\mF_P, \mF_C, \mF_M)\), producing the focal embeddings \(\tilde{\mathbf{X}}\). The global and focal features are then combined via element-wise addition and reduced via spatial pooling~\cite{m3d}. These visual tokens are then linearly projected into the language model's text embedding space:
\begin{align}
    \mathbf{T} = \big(\mathbf{X} + \tilde{\mathbf{X}}\big) \in \mathbb{R}^{K \times d_v}.
\end{align}
\begin{align} \label{eq:V}
    \mathbf{V} = \mathrm{Proj}(\mathrm{SpatialPooler}(\mathbf{T})).    
\end{align}

Finally, the projected tokens \(\mathbf{V}\) and a lesion-description query \(q\) are passed to the language model decoder to generate the localized findings $y$:
\begin{align}
    y = \text{LLMDecoder}(q, \mathbf{V}).
\end{align}

\subsection{Training stages}
We employ a three-stage training pipeline, where each stage optimizes the same objective in Eq.~(\ref{eq:loss}) while updating either partial or full model parameters. The entire procedure is first applied to our curated TotalSegmentator (TS) dataset for pretraining, where each sample is formatted as a question–answer pair: \textit{Question:} “What is the region highlighted by the mask?” \textit{Answer:} “The highlighted region is the $\langle$Region Class$\rangle$". The same three-stage pipeline is then repeated on PETARSeg-11K.

\textbf{Stage 1: Mask embedding alignment.}
We train only the projection head that maps PET/CT visual features into the language model’s embedding space in Eq.~(\ref{eq:V}). To ensure the learning focuses solely on this mapping, the mask-embedding weights are initialized to zero, and both the PET/CT encoder and the language model remain frozen. 

\textbf{Stage 2: Projector alignment.}
Next, we train the mask embedding module in Eq.~(\ref{eq:Z_M}) while freezing all other components. As shown in Figure~\ref{fig:overall} (right), the mask weights are trained to encode binary masks in a way that aligns with the underlying 3D anatomical and metabolic signals. This encourages the vision encoder to build meaningful correspondences between mask shapes and anatomical context.

\textbf{Stage 3: Full finetuning.}
Finally, we unfreeze the full architecture and perform end-to-end finetuning. In this stage, PET/CT features, mask-conditioned representations, and the language model are jointly optimized to generate accurate region-aware descriptions from the input mask.

\subsection{Training objective}
Given a training dataset \(\mathcal{D}\) of size \(D\), each example consists of PET--CT--mask visual triplets, a lesion-description query \(q\), and a target token sequence \(y = (y_1, \dots, y_N)\) of length \(N\). The model is trained to predict each token autoregressively, conditioned on the visual features and all previously generated tokens.
Formally, the objective is the autoregressive negative log-likelihood:
\begin{align} \label{eq:loss}
    \mathcal{L}(\mathcal{D}, \theta) = - \sum_{(\mathbf{V}, q, y) \sim \mathcal{D}} \sum_{i=1}^{N} \log p_\theta(y_i \mid \mathbf{V}, q, y_{<i}),
\end{align}
where \(\mathbf{V}\) denotes the visual tokens encoded from the PET--CT--mask encoder in Eq.~(\ref{eq:V}), and \(y_{<i}\) denotes all preceding tokens in the sequence. This is used for all stages of training. 
\section{Experiments and analysis}
\label{sec:experiments}

\begin{table*}[tbh]
\centering
\caption{Comparison of selected models on the PETARSeg-11k test set. The "(\textit{finetuned})" indicates the model was trained on PETARSeg-11k.}
\tiny
\resizebox{\textwidth}{!}{%
\begin{tabular}{l cccccccccc}
\toprule
\textbf{Model / Metric} & \textbf{BLEU} & \textbf{ROUGE-L} & \textbf{METEOR} & \textbf{CIDEr} & \textbf{BART} & \textbf{BERT} & \textbf{RaTE} & \textbf{GREEN} \\
\midrule
\addlinespace[1pt]
\rowcolor{sectiongray}\multicolumn{9}{l}{\textbf{2D VLMs}} \\[-0.3em]
\midrule
MedGemma-4B & 0.124 & 0.352 & 0.358 & 0.027 & -4.92 & 0.690 & 0.540 & 0.011 \\
HuatuoGPT-Vision-7B & 0.130 & 0.350 & 0.357 & 0.024 & -4.85 & 0.695 & 0.584 & 0.015 \\
InternVL3-8B & 0.137 & 0.355 & 0.356 & 0.035 & -4.72 & 0.694 & 0.614 & 0.030 \\
Qwen3-VL-8B & 0.375 & 0.343 & 0.364 & 0.051 & -4.74 & 0.690 & 0.612 & 0.022 \\
QoQ-Med & 0.364 & 0.317 & 0.368 & 0.025 & -4.64 & 0.680 & 0.505 & 0.006 \\
ViP-LLaVA & 0.064 & 0.301 & 0.308 & 0.009 & -5.96 & 0.651 & 0.503 & 0.006 \\
Qwen3-VL-8B (\textit{finetuned}) & 0.484 & 0.443 & 0.501 & 0.086 & -4.40 & 0.750 & 0.608 & 0.060 \\
MedGemma-4B (\textit{finetuned}) & 0.495 & 0.454 & 0.510 & 0.119 & -4.36 & 0.754 & 0.613 & 0.086 \\
\midrule
\rowcolor{sectiongray}\multicolumn{9}{l}{\textbf{3D VLMs}} \\[-0.3em]
\midrule
Med3DVLM & 0.004 & 0.080 & 0.066 & 0.005 & -5.70 & 0.511 & 0.285 & 0.002 \\
M3D & 0.327 & 0.276 & 0.323 & 0.013 & -5.32 & 0.634 & 0.505 & 0.000 \\
M3D-RAD & 0.343 & 0.300 & 0.340 & 0.016 & -5.23 & 0.658 & 0.518 & 0.003 \\
Reg2RG & 0.044 & 0.060 & 0.108 & 0.001 & -5.54 & 0.518 & 0.363 & 0.002 \\
Reg2RG (\textit{finetuned}) & 0.478 & 0.416 & 0.487 & 0.055 & -4.58 & 0.732 & 0.532 & 0.031 \\
M3D-RAD (\textit{finetuned}) & 0.485 & 0.446 & 0.501 & 0.132 & -4.34 & 0.750 & 0.627 & 0.071 \\
\textbf{PETAR-4B (\textit{Ours})} & \textbf{0.535} & \textbf{0.524} & \textbf{0.560} & \textbf{0.457} & \textbf{-4.00} & \textbf{0.795} & \textbf{0.713} & \textbf{0.257} \\
\bottomrule
\end{tabular}
}%
\label{tab:model_comparison}
\end{table*}

\begin{table*}[tbh]
\caption{Ablation study of our model showing the effect of each component on multiple evaluation metrics. TS=TotalSegmentator pretraining.}
\centering
\scriptsize
\resizebox{\textwidth}{!}{%
    \begin{tabular}{cccc|cccccccc}
        \toprule
        Mask & CT & Focal & TS & BLEU & ROUGE & METEOR & CIDEr & BERTScore & BARTScore & RaTEScore & GREEN \\
        \midrule
        \xmark & \xmark & \xmark & \xmark & 0.485 & 0.446 & 0.501 & 0.132 & 0.750 & -4.34 & 0.627 & 0.071 \\
        \cmark & \xmark & \xmark & \xmark & 0.480 & 0.445 & 0.498 & 0.137 & 0.748 & -4.33 & 0.626 & 0.088 \\
        \xmark & \cmark & \xmark & \xmark & 0.477 & 0.436 & 0.499 & 0.134 & 0.746 & -4.31 & 0.622 & 0.060 \\
        \xmark & \xmark & \cmark & \xmark & 0.528 & 0.518 & 0.550 & 0.397 & 0.787 & -4.05 & 0.698 & 0.226 \\
        \cmark & \xmark & \cmark & \xmark & 0.525 & 0.518 & 0.550 & 0.381 & 0.787 & -4.05 & 0.700 & 0.232 \\
        \xmark & \cmark & \cmark & \xmark & 0.517 & 0.507 & 0.540 & 0.428 & 0.784 & -4.08 & 0.587 & 0.234 \\
        \cmark & \cmark & \cmark & \xmark & 0.521 & 0.519 & 0.551 & 0.439 & 0.787 & -4.05 & 0.613 & 0.239 \\
        \cmark & \cmark & \cmark & \cmark & \textbf{0.535} & \textbf{0.524} & \textbf{0.560} & \textbf{0.457} & \textbf{0.795} & \textbf{-4.00} & \textbf{0.713} & \textbf{0.257} \\
        \bottomrule
    \end{tabular}
}%
\label{tab:ablation}
\end{table*}

\subsection{Experimental setup}

\textbf{Implementation}. We used the vision encoder and language model from M3D \cite{m3d}, a widely used medical VLM trained on large 3D medical image–text datasets. The vision encoder is a Vision Transformer with a 3D convolutional projection layer, and the language model is Phi3-4B \cite{phi3}. All experiments were done on 2 NVIDIA L40S GPUs. All stages were trained for 10 epochs, taking about 20 hours total. A complete list of hyperparameters are provided in the supplementary material.

\textbf{Test set}. As PETARSeg-11k is the first dataset of its kind for PET scans, evaluation was performed on a held-out set of 1175 test samples. We also manually curated a test set of size 32 from the autoPET \cite{autoPET} dataset for our reader evaluations. This dataset does not have corresponding reports and therefore automatic evaluation metrics do not apply. 

\textbf{Compared models}. We compared a variety of 2D and 3D VLMs, including both standard and prompt-aware variants. For 2D models, we used strong general-purpose and medical-specific baselines: InternVL3-8B \cite{internvl32025}, Qwen3-VL-8B-instruct \cite{qwen3}, HuatuoGPT-Vision-7B \cite{chen2024huatuogptvision}, MedGemma-4B \cite{sellergren2025medgemmatechnicalreport}, and QoQ-Med \cite{dai2025qoqmed}. The 2D prompt-aware model ViP-LLaVA \cite{vipllava} adds visual grounding with mask-conditioned prompting. For 3D models, we evaluated M3D \cite{m3d}, Med3DVLM \cite{xin2025med3dvlm} and M3D-RAD \cite{gai20253d-rad}, along with the 3D prompt-aware Reg2RG \cite{reg2rg}, which incorporates mask-guided conditioning. We finetuned representative 2D and 3D models on our dataset for fair comparison and validation. For 2D finetuning, we followed a similar strategy to \cite{pet2rep} and used 3 representative slices of the PET scan from the sagittal, coronal, and axial views to approximate 3D context. During inference stage for 2D models, these 3 slices were provided and the model was queried to produce findings for these images. During inference for 3D models, the entire volume was provided. 

\textbf{Metrics} We evaluated model predictions using 3 categories of metrics. N-gram overlap metrics (BLEU \cite{papineni2002bleu}, ROUGE \cite{lin2004rouge}, METEOR \cite{banerjee2005meteor}, CIDEr \cite{vedantam2015cider}), which measure precision, recall, and consensus-weighted word overlap; semantic metrics (BERTScore \cite{zhang2020bertscore}, BARTScore \cite{yuan2021bartscore}), which use contextual embeddings to measure semantic similarity; and language model–based metrics (RaTEScore \cite{zhao2024ratescore}, GREEN \cite{ostmeier-etal-2024-green}), which use large language models to assess clinical correctness, factual consistency, and reasoning.

\subsection{Main results}

We found that current VLMs, primarily trained on CT and radiography data, are limited in their ability to interpret PET data, as shown in \Cref{tab:model_comparison}. This indicates that the domain shift introduced by PET images and reporting styles is substantial. 3D VLMs like M3D-RAD and Med3DVLM showed limited captioning ability, with GREEN scores of just 0.002–0.03 and BERTScores below 0.66. Thus, our proposed dataset offers an important resource for bridging the performance of VLMs from CT to PET/CT.

Finetuning existing vision–language models on our PET dataset yielded consistent performance gains, particularly in clinically grounded metrics such as GREEN and RaTE, which assess factual and interpretive accuracy. For instance, Qwen3-VL-8B improves GREEN from 0.022 to 0.066 after fine-tuning, and M3D-RAD rises from 0.003 to 0.071. In contrast, our PETAR-4B model—designed with mask-aware conditioning and focal prompting—achieves substantial absolute improvements across all linguistic, semantic, and clinical metrics. As shown in \Cref{tab:model_comparison}, PETAR-4B surpassed the strongest 2D model (MedGemma) in BLEU-4 (0.535 vs 0.495), ROUGE-L (0.524 vs 0.454), and METEOR (0.560 vs 0.510). It also exceeded the top 3D model M3D-RAD (finetuned) on BERTScore (0.795 vs 0.750) and RaTEScore (0.713 vs 0.627), while achieving a higher GREEN score (0.257 vs 0.071). These results indicate that beyond data exposure, the architectural choices in PETAR-4B—particularly the use of mask-guided feature conditioning and region-focused prompting—are central to achieving clinically meaningful improvements in PET report generation.

\begin{figure*}[tbh]
  \centering
  \includegraphics[width=\linewidth]{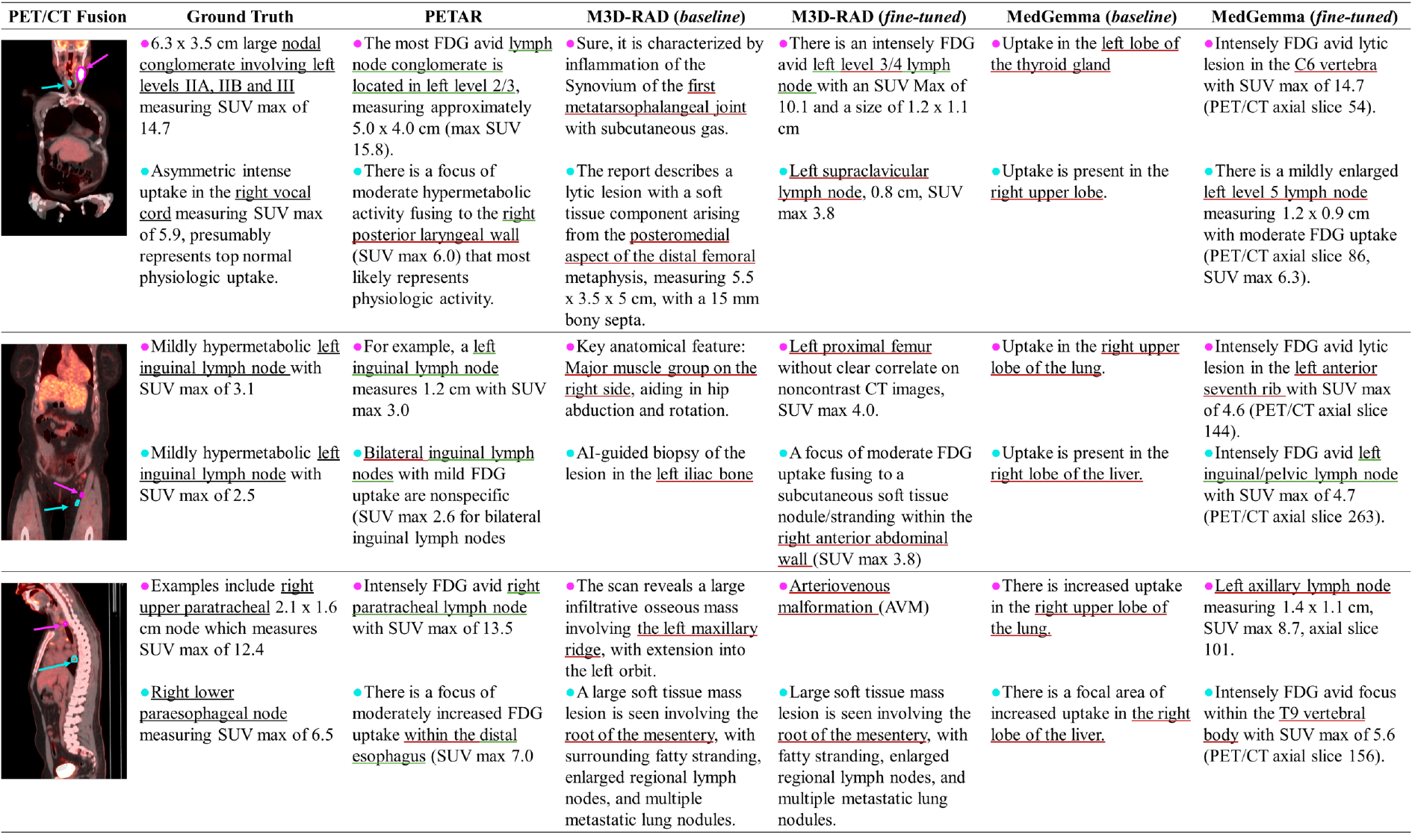}
  \caption{A comparison between different models. PETAR consistently produces anatomically correct descriptions. Prior to fine-tuning, both MedGemma and M3D-RAD produce highly inaccurate results. After fine-tuning, the models still tend to make localisation errors. Anatomical descriptors are underlined for ease of comparison (\underline{\textcolor{red}{red=incorrect}}, \underline{\textcolor{green!50!black}{green=correct}}). Note: quantitative measurements (lesion size, SUVmax) are hallucinated by the models but can be easily replaced with directly measured values using the input lesion masks.}
  \label{fig:examples}
\end{figure*}

Qualitative results are presented in \Cref{fig:examples}. These results illustrate the difficulty of the task, both in accurately describing the location and visual features of a lesion, as well as correctly interpreting it (e.g., "likely represents physiological activity"). The figure includes comparisons between PETAR and M3D-RAD before and after fine-tuning. In the zero-shot setting, M3D-RAD hallucinated irrelevant anatomy (e.g., describing the "maxillary ridge" instead of the correct paratracheal node). After fine-tuning on PETARSeg-11K (Fig. 4B), M3D-RAD showed improvement but still frequently misidentifies regions. For example, labeling uptake in the left inguinal node as the “left proximal femur.” In contrast, PETAR consistently aligned textual descriptions with real visual features and anatomical context, demonstrating superior grounding and clinical interpretability. The strong quantitative gains in CIDEr (+0.325) and GREEN (+0.186) further validate this qualitative advantage, indicating that PETAR-4B not only captures descriptive meaning but also generates clinically meaningful descriptions. MedGemma, shows similar performance trends as M3D-RAD. 

\subsection{Human evaluation}

Though automated metrics are widely used to evaluate medical report generation, it is unclear how well they reflect what clinicians actually value. Many metrics focus on word overlap rather than semantic meaning, so they may penalize valid paraphrases or reward text that sounds correct but is factually wrong. Even newer, language model–based metrics such as GREEN, which aim to capture meaning, can still overrate text that seems plausible but introduces unsupported details.

To assess our model's performance from an expert perspective, we conducted a human evaluation with five board-certified nuclear medicine physicians. They reviewed 116 pairs of real (i.e., physician-generated) and PETAR-generated lesion descriptions, blinded to the source of the descriptions, and rated each on a standardized 5-point scale for anatomical accuracy, interpretation correctness, and clinical usefulness. We excluded errors related to quantitative measurements (e.g., lesion size) since those values are sometimes hallucinated by the model, but can be directly derived from the input masks, so they are not the focus of our evaluation. They also evaluated 32 instances of the autoPET dataset for testing on unseen distributions. To the best of our knowledge, this is the first human study of PET report generation by vision-language models.

\Cref{tab:human_eval} shows that PETAR-4B produces clinically useful findings, with anatomical, interpretation, and utility scores all in the 3.7–3.9 range compared to human scores of 4.3–4.4. Furthermore, the physicians preferred, or considered equal, the model generated findings in $\approx60\%$ of cases (69/116). PETAR-4B achieved similarly strong scores on the external AutoPET dataset, indicating consistent performance for out-of-distribution data. 

 To explore which evaluation metrics best align with human judgment for PET report generation, we used the ratings from the reader study and analyzed how existing automated metrics correlate with human scores.  Results are shown in \Cref{tab:human_corr}. Metrics that capture semantic and contextual similarity showed markedly stronger correlation with radiologist preferences compared to traditional n-gram–based scores like BLEU. This finding underscores the need for evaluation standards that better reflect clinical reasoning rather than surface-level text similarity. GREEN provided the best indicator of PET report quality ($\rho$=0.59).

\begin{table}[t]
\centering
\caption{Spearman’s correlation between automated metrics and human evaluation scores. Higher correlation indicates stronger alignment with expert judgment.}
\begin{tabular}{lc}
\toprule
\textbf{Metric} & \textbf{Spearman’s $\rho$} \\
\midrule
GREEN      & 0.592 \\
RaTEScore  & 0.550 \\
BERTScore  & 0.511 \\
ROUGE      & 0.471 \\
METEOR     & 0.438 \\
CIDEr      & 0.421 \\
BLEU       & 0.214 \\
BARTScore  & 0.168 \\
\bottomrule
\end{tabular}
\label{tab:human_corr}
\end{table}


\subsection{Ablation studies}

To evaluate the contribution of key design components in PETAR-4B, we performed ablation experiments on the mask input, CT modality, focal prompt, and TotalSegmentator pretraining (\Cref{tab:ablation}).  Removing any of PETAR's modules led to a measurable decrease in performance across all metrics. The mask input provided crucial spatial grounding, substantially improving lesion localization and factual accuracy. Adding the CT modality enhanced anatomical coherence and contextual understanding, which translates to higher BERTScore and GREEN. The focal prompt had the strongest impact overall, sharpening attention on fine-grained details that might otherwise be lost.  TotalSegmentator pretraining also provided a measurable boost in performance. By combining all components, PETAR-4B achieved the best performance, indicating PET/CT reasoning benefits from the synergy between volumetric grounding and lesion-focused prompting.

\subsection{Limitations and future work}

A feature of our method is the requirement of mask inputs to achieve the best performance. Currently, these would need to come from physicians. However, this could easily fit into a typical physician reading workflow, where each lesion could be contoured by a physician using single-mouse-click segmentation methods (MIM's PETEdge+, which already has wide clinical adoption), and then the report findings would be drafted using PETAR. However, a fully automated pipeline is possible, wherein state-of-the-art lesion segmentation algorithms first identify and segment potential findings, which serve as inputs to PETAR. Substantial work has already been published on automated PET lesion detection and segmentation \cite{Rokuss_2025_CVPR, LASnet, TMTVNet}. In future work, we will explore ways to build this pipeline using such models, and how to best convert lesion descriptions into templated PET reports. 

\begin{table}[tbh]
\centering
\caption{Human evaluation of \textbf{PETAR-4B} on an internal  test set and an external AutoPET dataset. Five blinded nuclear medicine physicians compared PETAR findings with original reports.}
\label{tab:human_eval}
\resizebox{\columnwidth}{!}{
\begin{tabular}{lccc}
\toprule
\multirow{2}{*}{\textbf{Metric}} & \multicolumn{2}{c}{\textbf{Internal Reader Study}} & \textbf{AutoPET (External)} \\
\cmidrule(lr){2-3} \cmidrule(lr){4-4}
 & \textbf{Human} & \textbf{PETAR-4B} & \textbf{PETAR-4B} \\
\midrule
Anatomical score     & 4.4 $\pm$ 0.9 & 3.9 $\pm$ 1.2 & 3.8 $\pm$ 1.1 \\
Interpretation score & 4.4 $\pm$ 0.8 & 3.9 $\pm$ 1.1 & 4.0 $\pm$ 1.0 \\
Utility score        & 4.3 $\pm$ 0.9 & 3.7 $\pm$ 1.1 & 3.8 $\pm$ 1.1 \\
\midrule
PETAR preferred (\# cases)& \multicolumn{2}{c}{15} & -- \\
Preference tied (\# cases)   & \multicolumn{2}{c}{54} & -- \\
PETAR not preferred (\# cases)& \multicolumn{2}{c}{47} & -- \\
\bottomrule
\end{tabular}
}
\end{table}
\section{Conclusion}
\label{sec:conclusion}

We introduced PETAR, a 3D mask-aware vision–language framework for PET/CT report generation, along with PETARSeg-11k, the first large-scale dataset linking lesion-level PET/CT findings to descriptive clinical text. By combining metabolic (PET) and structural (CT) cues through mask-guided volumetric encoding and focal prompting, our model captures fine-grained details of small, spatially-dispersed lesions often lost by other models. PETAR sets a new state-of-the-art performance across linguistic, semantic, and clinically grounded metrics. Our evaluations, including a human study with five board-certified physicians, confirmed the model’s ability to produce accurate and clinically useful findings that show strong alignment with expert judgment. 

\section{Acknowledgments}

Research reported in this publication was supported by the National Institute Of Biomedical Imaging And Bioengineering of the National Institutes of Health under Award Number R01EB033782. The content is solely the responsibility of the authors and does not necessarily represent the official views of the National Institutes of Health.
{
    \small
    \bibliographystyle{ieeenat_fullname}
    \bibliography{main}
}

\clearpage
\setcounter{page}{1}
\maketitlesupplementary

\section{Dataset information}
\label{sec:supp-dataset}
\subsection{Collection}
PET/CT images were collected from patient scans conducted between 2010-2013, using scanners from GE Healthcare, at the University of Wisconsin-Madison hospital. Of the 5126 unique exams, 4798 are FDG, 178 are DOTATATE, 106 are  [18F]Fluciclovine, and 44 are [18F]DCFPyL. Images were originally formatted in DICOM and converted into NIfTI. PET images were converted to units of SUV. Reports were formatted in csv. DICOM images were deidentified using Clinical Trial Processor, and reports were deidentified using NLM Scrubber \cite{NLM_Scrubber}. 

The autoPET dataset is already deidentified, processed, and made available through the works of Gatidis \etal \cite{autoPET}. Note that the CT images for the AutoPET dataset were primarily contrast-enhanced CTs (i.e., iodine contrast), whereas our internal dataset consisted primarily of non-contrast-enhanced CTs. 
\subsection{Statistics}
We collected statistics for our dataset regarding age, sex, and disease type. This is summarized in \Cref{tab:summary}, \Cref{tab:indication}, and \Cref{tab:cancer}.

\begin{table}[t]
\centering
\caption{Cohort summary statistics.}
\small
\begin{tabular}{l c}
\toprule
\textbf{Summary Statistics} & \textbf{Value (\%)} \\
\midrule
Average Age (years) & 62.2 \\
Male         & 53.2 \\
Female       & 46.9 \\
\bottomrule
\end{tabular}
\label{tab:summary}
\end{table}

\begin{table}[t]
\centering
\caption{Indication frequency with low-prevalence categories grouped into ``Others'' (percent only).}
\small
\begin{tabular}{l c}
\toprule
\textbf{Indication} & \textbf{Percent (\%)} \\
\midrule
Restaging                          & 39.7 \\
Initial staging                    & 29.6 \\
Treatment response assessment      & 13.8 \\
Metastatic workup                  & 5.2 \\
Suspected recurrence               & 3.2 \\
Others                  & 8.5 \\
\bottomrule
\end{tabular}
\label{tab:indication}
\end{table}

\begin{table}[t]
\centering
\caption{Cancer type frequency with rare categories grouped into ``Others'' (percent only).}
\small
\begin{tabular}{l c}
\toprule
\textbf{Cancer Type} & \textbf{Percent (\%)} \\
\midrule
Lymphoma                                 & 22.8 \\
Lung                                     & 22.5 \\
Head and neck                            & 11.5 \\
Breast                                   & 7.6 \\
Esophageal                               & 4.4 \\
Melanoma                                 & 4.2 \\
Gynecologic (ovarian/cervical/endometrial) & 3.6 \\
Colorectal                               & 3.5 \\
Neuroendocrine tumor                     & 3.3 \\
Gynecologic cancer                       & 3.1 \\
Prostate                                 & 2.7 \\
Others                         & 20.8 \\
\bottomrule
\end{tabular}
\label{tab:cancer}
\end{table}

\subsection{Formatting}
We used the following prompt for the LLM to structure the data in the format highlighted our dataset construction section. 

\begin{quote}
    "Given the following PET/CT finding:

    \{Findings\}
    
    Extract and format the information in this exact structure:
    
    Region: Broad body section (e.g., Head, Neck, Chest, Abdomen, Pelvis etc.) 
    
    Organ: The nearest major organ or system involved (e.g., liver, pancreas, kidney, lung, bone etc.) 
    
    Anatomic Subsite: Specific substructure or precise location described (e.g., peripancreatic, left adrenal, posterior to left kidney, upper right gluteal etc.) 
    
    SUV Max: Numerical SUV max value mentioned 
    
    Axial Slice: Axial slice number mentioned
    
    Findings: Copy the original sentence exactly as written Rules: If any field is not explicitly stated, write ``N/A''. Do not add interpretations beyond the sentence. Enclose the final output strictly between $<$extract$>$ and $<$/extract$>$ tags."

\end{quote}

\section{Further implementation details}
\label{sec:supp-implementation}
\subsection{Training environment}

We use the MONAI library \cite{monai} for 3D volumetric data processing, and PyTorch \cite{paszke2019pytorch} together with the HuggingFace \texttt{transformers} and \texttt{accelerate} libraries \cite{wolf-etal-2020-transformers} for model creation, optimization, and distributed training. All experiments are launched using \texttt{accelerate launch} and leverage LoRA-based parameter-efficient fine-tuning for language models \cite{lora}. Mixed-precision training was enabled using \texttt{bfloat16} \cite{kalamkar2019studybfloat16deeplearning}.

\subsection{Hyperparameters}
We fine-tuned the combination of our modified image encoder and the language component of M3D \cite{m3d} using LoRA adapters applied to all linear layers in the language backbone. Evaluation and checkpointing are performed at fixed step intervals. All hyperparameters used during training are listed in Table~\ref{tab:train_hyperparameters}.

\begin{table}[t]
\centering
\caption{Hyperparameters used for LoRA-based fine-tuning of PETAR-4B}
\small
\begin{tabular}{p{0.5\columnwidth} p{0.4\columnwidth}}
\toprule
\textbf{Hyperparameter} & \textbf{Value} \\
\midrule
Mixed precision & bf16 \\
Model max length & 512 \\
Training epochs & 5 \\
Per-device batch size & 2 \\
Gradient accumulation & 1 \\
Learning rate & 5e-5 \\
Weight decay & 0.0 \\
Warmup ratio & 0.03 \\
LR scheduler & cosine \\
Gradient checkpointing & False \\
Evaluation strategy & steps \\
Evaluation steps & 0.2 \\
Eval accumulation steps & 1 \\
Dataloader workers & 8 \\
Pin memory & True \\
\bottomrule
\end{tabular}
\label{tab:train_hyperparameters}
\end{table}

\begin{table}[t]
\centering
\caption{Comparison of interpretation options for patient 1.}
\small
\begin{tabularx}{\columnwidth}{l l l X}
\toprule
\textbf{Patient ID} & \textbf{ROI ID} & \textbf{Preference} & \textbf{Description} \\
\midrule
Patient 1 & Mask1 & Option 1 &
There is a small, intensely FDG avid focus in the right iliac fossa with SUV max of [\#] (PET/CT axial slice [\#]) that is not seen on CT. \\
\midrule
Patient 1 & Mask1 & Option 2 &
Intensely radiotracer avid nodular focus ([\#] x [\#] cm) in close proximity to the surgical clips, measuring SUV max of [\#] correlates with the recently CT characterized nodularity and is most suspicious for recurrent disease. \\
\bottomrule
\end{tabularx}
\label{tab:patient_interpretation_table}
\end{table}

\section{AutoPET examples}
\begin{figure}[t]
  \centering
  \includegraphics[width=\linewidth]{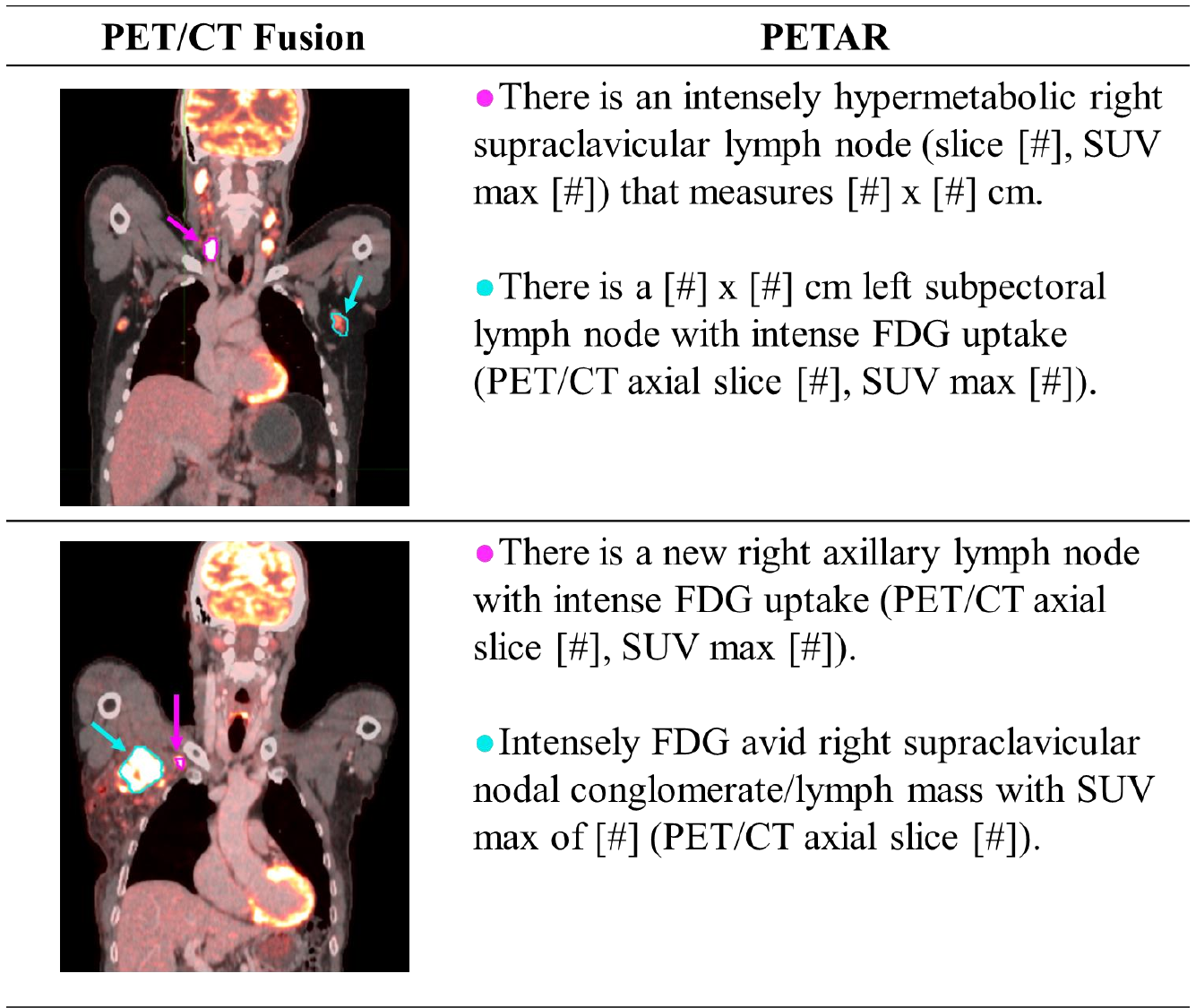}
  \caption{PETAR-4B predictions for examples from the autoPET dataset.}
  \label{fig:autopet_examples}
\end{figure}

In \Cref{fig:autopet_examples}, we show examples of findings produced by PETAR-4B on the autoPET data. 

\begin{table}[t]
\centering
\caption{Anatomical accuracy rubric.}
\small
\begin{tabular}{p{0.8cm} p{0.75\columnwidth}}
\toprule
\textbf{Score} & \textbf{Description} \\
\midrule
1 & Completely incorrect. Wrong organ/system, wrong side, or irrelevant location. \\
\midrule
2 & Mostly incorrect. Correct general region but wrong substructure. \\
\midrule
3 & Partially correct. Correct organ/region but poor extent or boundary mismatch. \\
\midrule
4 & Nearly correct. Correct location with minor boundary issues. \\
\midrule
5 & Completely correct. Precise location and extent. \\
\bottomrule
\end{tabular}
\label{tab:anatomical_accuracy}
\end{table}

\section{Human evaluation details}
\label{sec:supp-humaneval}
\subsection{Evaluation dataset preparations}
We provided physicians with DICOM PET/CT images, with lesion contour overlays, in MIM Encore software. In the provided text, we replaced all numerics to [\#] as these are hallucinated and can easily be extracted from the input contour. For every pair of ground truth and prediction, we shuffled the order for all examples to mitigate any bias in scoring. An example of what the physicians saw for scoring is shown in \Cref{tab:patient_interpretation_table}.

\begin{table*}[t]
\centering
\small
\begin{minipage}[t]{0.48\textwidth}
\centering
\captionof{table}{Interpretation accuracy rubric.}
\begin{tabular}{p{0.8cm} p{0.75\linewidth}}
\toprule
\textbf{Score} & \textbf{Description} \\
\midrule
1 & Completely incorrect; misleading or dangerous. \\
\midrule
2 & Mostly incorrect; plausible but clinically misleading. \\
\midrule
3 & Partially correct; right lesion but wrong qualifier. \\
\midrule
4 & Nearly correct; accurate meaning but missing detail. \\
\midrule
5 & Completely correct; matches ground truth interpretation. \\
\bottomrule
\end{tabular}
\label{tab:interpretation_accuracy}
\end{minipage}
\hfill
\begin{minipage}[t]{0.48\textwidth}
\centering
\captionof{table}{Overall utility rubric.}
\begin{tabular}{p{0.8cm} p{0.75\linewidth}}
\toprule
\textbf{Score} & \textbf{Description} \\
\midrule
1 & No utility; unusable or harmful. \\
\midrule
2 & Minimal utility; major edits required. \\
\midrule
3 & Moderate utility; usable after moderate edits. \\
\midrule
4 & High utility; only minor edits required. \\
\midrule
5 & Fully useful; clinically usable as written. \\
\bottomrule
\end{tabular}
\label{tab:overall_utility}
\end{minipage}

\end{table*}

\subsection{Guidance on scoring}
We reproduce the scoring guidelines provided to the physicians below in \Cref{tab:anatomical_accuracy}, \Cref{tab:interpretation_accuracy}, \Cref{tab:overall_utility}. As the autoPET dataset does not have corresponding reports, we do not report preference scores or automated metrics as these require a reference ground truth.
\end{document}